\documentclass[english,a4paper,11pt]{article}
\usepackage[top = 2cm, bottom = 2cm,left = 2cm, right = 2cm]{geometry}
\usepackage[T1]{fontenc}
\usepackage[latin9, utf8]{inputenc}
\usepackage{bm}
\usepackage{amsmath}
\usepackage{amssymb}
\usepackage{nccmath}
\usepackage{graphicx}

\usepackage{caption}
\usepackage[lofdepth,lotdepth]{subfig}

\usepackage[linesnumbered,ruled,vlined]{algorithm2e}
\usepackage{mwe}
\usepackage{babel}
\usepackage{multirow}
\usepackage{makecell}
\usepackage{mathrsfs}
\usepackage[affil-it]{authblk}
\usepackage[colorlinks=true, allcolors=blue]{hyperref}
\usepackage{titles}
\usepackage{float} 
\usepackage{siunitx}

\usepackage{indentfirst}

\usepackage{abstract}

\title{End-to-end Wind Turbine Wake Modelling with Deep Graph Representation Learning}

\author[1]{\small Siyi Li}
\author[1]{Mingrui Zhang}
\author[1]{\small Matthew D. Piggott\thanks{Corresponding author.\newline E-mail address: m.d.piggott@imperial.ac.uk}}
\date{}

\affil[1]{Department of Earth Science and Engineering,
Imperial College London, London, SW7 2AZ, UK}

\begin{document}
\maketitle

\begin{abstract}
Wind turbine wake modelling is of crucial importance to accurate resource assessment, to layout optimisation, and to the operational control of wind farms. This work proposes a surrogate model for the representation of wind turbine wakes based on a state-of-the-art graph representation learning method termed a graph neural network. The proposed end-to-end deep learning model operates directly on unstructured meshes and has been validated against high-fidelity data, demonstrating its ability to rapidly make accurate 3D flow field predictions for various inlet conditions and turbine yaw angles. The specific graph neural network model employed here is shown to generalise well to unseen data and is less sensitive to over-smoothing compared to common graph neural networks. A case study based upon a real world wind farm further demonstrates the capability of the proposed approach to predict farm scale power generation. Moreover, the proposed graph neural network framework is flexible and highly generic and as formulated here can be applied to any steady state computational fluid dynamics simulations on unstructured meshes. 

\end{abstract}

\section{Introduction} \label{introduction}
As one of the cleanest and most sustainable sources of renewable energy, wind energy has been undergoing rapid and unabated expansion worldwide.
As the capacity of wind turbine farms increases, through the potentially closer clustering of increasing numbers of larger turbines to most efficiently exploit the available wind energy resource, it is inevitable that downstream turbines will at some times be operating within the full or partial wakes of upstream turbines. This can lead to reduced power generation as well as increased structural loads. Consequently, wind turbine wake modelling has been widely considered as one of the most crucial aspects of the optimal design and operational control of wind farms, see \cite{PiggottReview2022} and the references therein. 

Wake models across different levels of fidelity have been thoroughly studied by researchers over the years. Analytical models including the Jensen model \cite{jensen}, the Larsen model \cite{larsen} and the Gaussian wake model \cite{gauss} are commonly implemented in industrial standard software such as FLORIS \cite{floris}, thanks to their very rapid execution speed, however their accuracy is consequently limited. In comparison, higher fidelity models based on computational fluid dynamics (CFD) simulations, such as Reynolds-Averaged Navier-Stokes (RANS) or Large Eddy Simulation (LES), can provide more accurate flow field predictions but at significantly higher computational cost and execution time, hampering their value for rapid resource assessment, and as part of iterative design optimisation and control tools. For instance, the computing time required by RANS modelling for the simulation of a wind farm tends to be in the order of several CPU hours, whereas LES simulations could take days of distributed computation on hundreds of processors \cite{les}. 

One possible approach to retain high accuracy in wake predictions while simultaneously maintaining short computation times is through the utilisation of deep learning algorithms trained on high-fidelity CFD data. The work presented here aims to develop a novel data-driven wake model that is based on machine learning and high-fidelity CFD simulations. In particular, this work utilises a graph representation learning method termed a graph neural network (GNN), which is based on a nascent deep learning research area operating on graph structured data. By operating directly on unstructured CFD meshes, the GNN approach eliminates the need for the training data to be interpolated to a uniform grid, as is commonly performed. It can therefore better preserve flow details through the accommodation of the different spatial resolutions across the simulation domain that are often beneficial for the CFD study of multi-scale fluid dynamical processes. A well-trained GNN on high-fidelity flow field data is thus able to capture the entirety of the primary characteristics of the wake flow structure, which cannot be achieved by analytical wake models, while maintaining competitive evaluation speeds.


\noindent \newline The primary novelties and contributions of this work can be summarised as follows:

\begin{enumerate}
    \item A novel wind turbine wake model based on graph representation learning was developed and trained on RANS CFD data. The use of graph neural networks is of particular significance as many leading high-fidelity CFD solvers seek to increase their efficiency in the simulation of complex, multi-scale problems through the use of unstructured meshes, block-structured meshes, or some other non-regular discretisation of the spatial domain. This is in contrast to most of the prior works on deep learning methods for wind turbine wake modelling which have typically operated on uniform grids.
    \item To the best of the authors' knowledge, this work is also one of the first attempts to leverage graph neural networks in the modelling of relatively large-scale, 3D CFD data comprising of the order of hundreds of thousands of graph vertices. The nature of this task entails the construction of deep graph neural networks, which are known to suffer from over-smoothing. To this end, this work explored the usage of various different graph neural network models and architectures, and conducted extensive experimentation, before adopting the {\em GraphSAGE} (Graph SAmple and aggreGatE) model which has great scalability with large data sizes. Moreover, the deep GraphSAGE neural network was further improved by adding a {\em jumping knowledge layer} and {\em layer-wise} as well as {\em initial residual connections}. 
    \item In order to train the deep learning surrogate model, RANS-based wake data was generated using the {\em generalised actuator disk (GAD) model} coupled with the CFD solver package {\em OpenFOAM}. The GAD model's ability to simulate turbine wakes and turbine wake interactions were further validated against wind tunnel tests performed at the Norwegian University of Science and Technology (NTNU).
    Also developed within the proposed framework was the ability to convert OpenFOAM based meshes to graph data structures that are compatible as input to GNNs. The performance of the GNN model in predicting single turbine wakes was extensively tested at varying levels of inlet velocities, turbulence intensities as well as turbine yaw angles. The resulting relative accuracy in predicting flow velocity on data unseen during training reached a median of 99.71$\%$, with each prediction capable of being made within 15 milliseconds. 
    \item The ability of the proposed geometric deep learning approach to model wind farms was further tested on a real-scale case study based on Sweden's Lillgrund offshore wind farm, by superimposing multiple individually calculated turbine wakes. The results showed that the proposed model can accurately predict the generated power in comparison to both full CFD simulations of the farm as well as direct observation data. In addition, the geometric deep learning surrogate model was able to simulate wind farms within seconds compared to many CPU hours of parallel computing which might be needed for a RANS simulation.
\end{enumerate}

The remainder of this article is organised as follows: A brief overview of closely related research works is given in Section \ref{related_work}. The numerical CFD model used to simulate wind turbine wakes for training data generation is introduced in Section \ref{numerical_model}, along with model validation studies against experimental data. The proposed graph representation learning based surrogate for wind turbine wake modelling is detailed in Section \ref{graph_representation_learning}. In addition to the preliminaries of graph neural networks, a series of training experiments on graph neural network architectures and a case study on the Lillgrund offshore wind farm. Conclusions and potential future plans for this work are detailed in Section \ref{finale}.
\section{Related Work}
\label{related_work}
This work is related to various previous works across different disciplines, such as machine learning, deep learning, general fluid dynamics as well as the specific application of wind turbine wake modelling. This section provides a brief overview of these diverse connections. 

\subsection*{Machine learning and CFD}
With the rapid advancements in computational power and an explosion of available data from experiments, simulations and historical records, machine learning has seen an increasing level of success in many scientific disciplines, including in computational fluid dynamics. For example, Thuerey et al. \cite{thurey_rans} explored the use of U-Net convolutional neural network (CNN) architecture to predict velocity and pressure around airfoils of different shapes based on RANS solutions. Guo et al. \cite{autodesk_CNN} trained CNNs to predict steady laminar flow past a range of 2D and 3D objects. More recently, geometric deep learning or deep learning on graphs has seen some limited applications in CFD. Belbute-Peres et al. \cite{belbute_peres} embedded a differentiable PDE solver as an implicit layer in a graph convolutional network framework and trained an end-to-end deep learning model with improved generalisation capabilities. Pfaff et al. \cite{meshgraphnet} developed a graph neural network based architecture for learning time-dependent physical simulations on meshes. Lino et al. \cite{lino_2} further considered message-passing at multiple scales of resolution and incorporated rotational equivariance into the graph neural network model. Suk et al. \cite{mesh_artery} trained gauge equivariant mesh convolutional networks to predict wall shear stresses over surface meshes that represented artery models. The work presented here builds on this area of research by modelling a large-scale, 3D CFD dataset on meshes with deep graph neural networks.

\subsection*{Deep learning based surrogate modelling of wind turbine wakes}

Despite the enormous potential for data-driven machine learning based wake models, there have been relatively few studies on the surrogate modelling of wind turbine wakes with machine learning methods. Ti et al. \cite{wake_mlp} used machine learning methods to predict turbine wake fields by interpolating actuator disk with rotation (ADM-R) RANS-based simulation data onto a uniform grid, before splitting it into 2000 partitions and training a multilayer perceptron (MLP) on each partition. Li et al. \cite{wake_CNN} developed a 2D dynamic wake model using bilateral CNN trained on high-fidelity LES data. Zhang et al. \cite{wake_GANs} also successfully trained a convolutional conditional generative adversarial neural network on LES data that could provide accurate real-time 2D wake predictions. The work presented here further contributes to the data-driven wake modelling research area by creating a 3D wake model with state-of-the-art geometric deep learning methods that is capable of predicting both velocity and turbulent kinetic energy (TKE) flow fields while also taking into consideration turbine yaw induced wake steering effects.
\section{Numerical Model}
\label{numerical_model}
This section provides a brief overview of the generalised actuator disk method as well as model validation with the ``blind test'' wind tunnel experiments performed at the Norwegian University of Science and Technology. 
\subsection{Generalised actuator disk (GAD) model}

\begin{figure}[H]
    \centering
    \includegraphics[width=\textwidth]{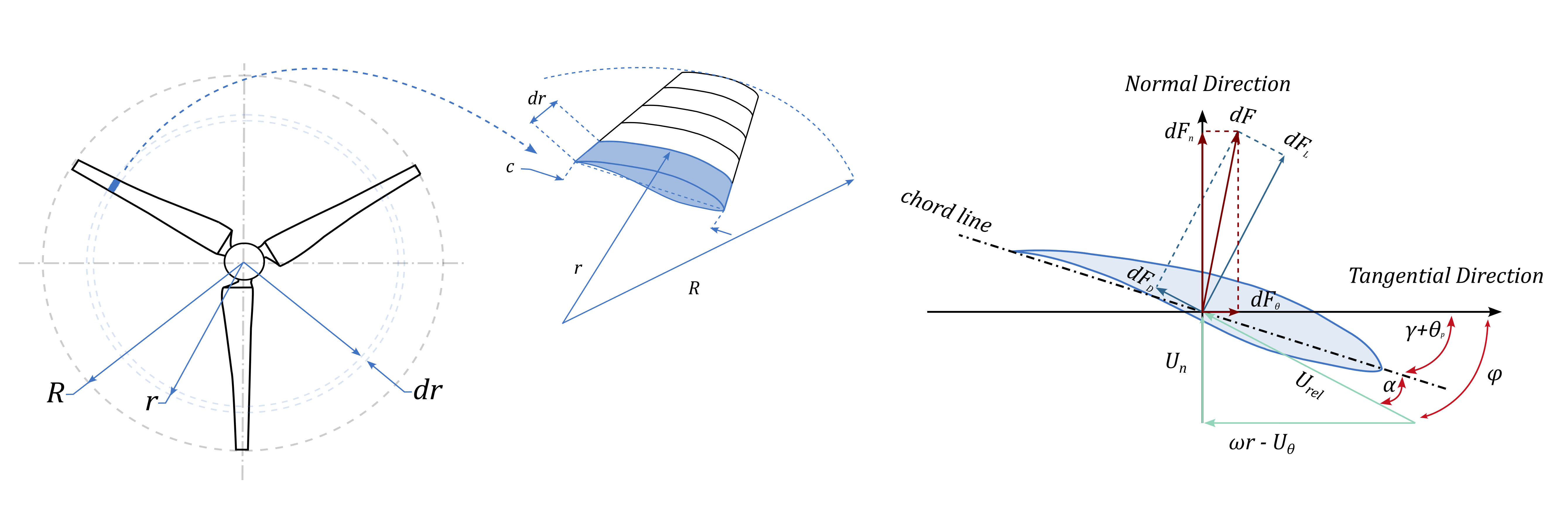}
    \caption{Blade element momentum discretisation of the rotor blades and the forces exerted on a single blade. $R$ stands for wind turbine radius and $r$ represents distance from a blade element to the centre of the wind turbine in the span-wise direction. The remaining symbols are defined in the main text.}
    \label{fig:BEM}
\end{figure}

The GAD model parameterises the wind turbine rotor as a virtual permeable disk, where the rotor blades are divided in the radial direction into various sections. The lift and drag forces exerted on the rotor blades by the fluid flow are computed with blade element momentum (BEM) theory on each control volume, summed up and then multiplied by the number of blades to be incorporated into the Navier-Stokes momentum equation as an additional source term. The forces acting on a blade element with length $dr$ in the span-wise direction can be written as follows:
\begin{fleqn}
    \begin{equation}
        \begin{aligned}
            &dF_D = \frac{1}{2} \rho c |U_{\text{rel}}|^2 C_D \, dr,& \\
            &dF_L = \frac{1}{2} \rho c |U_{\text{rel}}|^2 C_L\,dr,& \\
            &dF_n = F_L \; \cos \; \varphi + F_D \; \sin \; \varphi,& \\
            &dF_{\theta} = F_L \; \sin \; \varphi - F_D \; \cos \; \varphi,&
        \end{aligned}
    \end{equation}
\end{fleqn}
where $dF_L$, $dF_D$, $dF_n$ and $dF_{\theta}$ represent respectively the lift, drag, normal and tangential forces on a blade element, $U_{\text{rel}}$ denotes the relative wind velocity, $c$ is the chord length which varies in the radial direction and $C_L$ and $C_D$ are the lift and drag coefficients respectively, and are dependent on the local angle of attack $\alpha$ and the Reynolds number $Re$. $\varphi$ is the flow inclination angle given by:
\begin{fleqn}
    \begin{equation}
        \varphi = \tan^{-1}\left(\frac{\omega r - U_{\theta}}{U_n}\right) = \gamma + \theta_p + \alpha,
    \end{equation}
\end{fleqn}
where $\omega$ denotes the turbine angular velocity, $U_{\theta}$ and $U_n$ are the tangential and normal velocity, $\gamma$, $\theta_p$ and $\alpha$ are the blade twist angle, blade pitch angle and angle of attack respectively. The momentum equation source term can be computed as:
\begin{fleqn}
    \begin{equation}
        \begin{aligned}
            &S_i = S_n + S_{\theta}, \\
            &S_n = N \, dF_n \, \hat{v}_n =  \frac{1}{2} \rho N c |U_{\text{rel}}|^2(C_L \; \cos \; \varphi + C_D \; \sin \; \varphi) \, dr \; \hat{v}_n,\\
            &S_{\theta} = N \, dF_{\theta} \, \hat{v}_{\theta} =  \frac{1}{2} \rho N c |U_{\text{rel}}|^2(C_L \; \sin \; \varphi - C_D \; \cos \; \varphi) \, dr \; \hat{v}_n,
        \end{aligned}
    \end{equation}
\end{fleqn}
where $N$ is the number of turbine blades, $\hat{v}_n$ and $\hat{v}_{\theta}$ are the unit vector in the normal and tangential directions respectively. The nacelle effect was modelled in a similar way here as a blade element, but with a constant drag coefficient with $C_{D, \text{nacelle}} = 1$ and zero lift ($C_{L, \text{nacelle}} = 0$). An illustration of the GAD discretisation scheme and the forces applied on a single rotor blade is shown in Figure \ref{fig:BEM}.

The thrust, torque and power can then be calculated by integrating the forces over the virtual disk as follows:

\begin{fleqn}
    \begin{equation}
        \begin{aligned}
            &T = \sum_{i} \rho V_i S_{n,i},  \\
            &Q = \sum_{i} \rho V_i r_i S_{\theta, i},  \\
            &P = \sum_{i} \rho V_i r_i \omega S_{\theta, i},
        \end{aligned}
    \end{equation}
\end{fleqn}
where $T$, $Q$, and $P$ represent thrust, torque and power respectively, $i$ is the cell index of the list of cells that lie within the discretised actuator disk region, and $V_i$ is the cell volume
associated with cell $i$. The numerical modelling framework was implemented here in the {\em OpenFOAM} CFD package \cite{openfoam} using a modified version of the {\em GAD-CFD} code developed in \cite{GAD_paper} and coupled with the steady-state {\em SimpleFoam} solver. In-depth theoretical background and implementation details of the GAD model can be found in numerous recent studies \cite{GAD_theory, GAD_paper_2, GAD_paper_3}.

\subsection{Model validation}
The GAD model has been extensively validated against various experimental studies \cite{GAD_paper}; for instance, against a series of tidal turbine experiments at the French Research Institute for Exploitation of the Sea (IFREMER) by Mycek et al. \cite{Mycek}, experimental work of Selig et al. at the National Renewable Energy Laboratory (NREL) with a full-scale NREL phase III horizontal axis wind turbine \cite{NREL_phase_3}, as well as a large-scale experiment with multiple tidal turbines at the FloWave ocean energy research facility \cite{GAD_valid}. This work further validated it against wind tunnel experiments conducted by the Norwegian University of Science and Technology (NTNU) \cite{BT1_exp, BT2_exp}, referred to as NTNU ``blind test'' (BT1) and ``blind test 2'' (BT2).

\begin{figure}[H]
    \centering
    \subfloat[][Schematic representation of NTNU BT1.]{
    \includegraphics[width=0.45\textwidth]{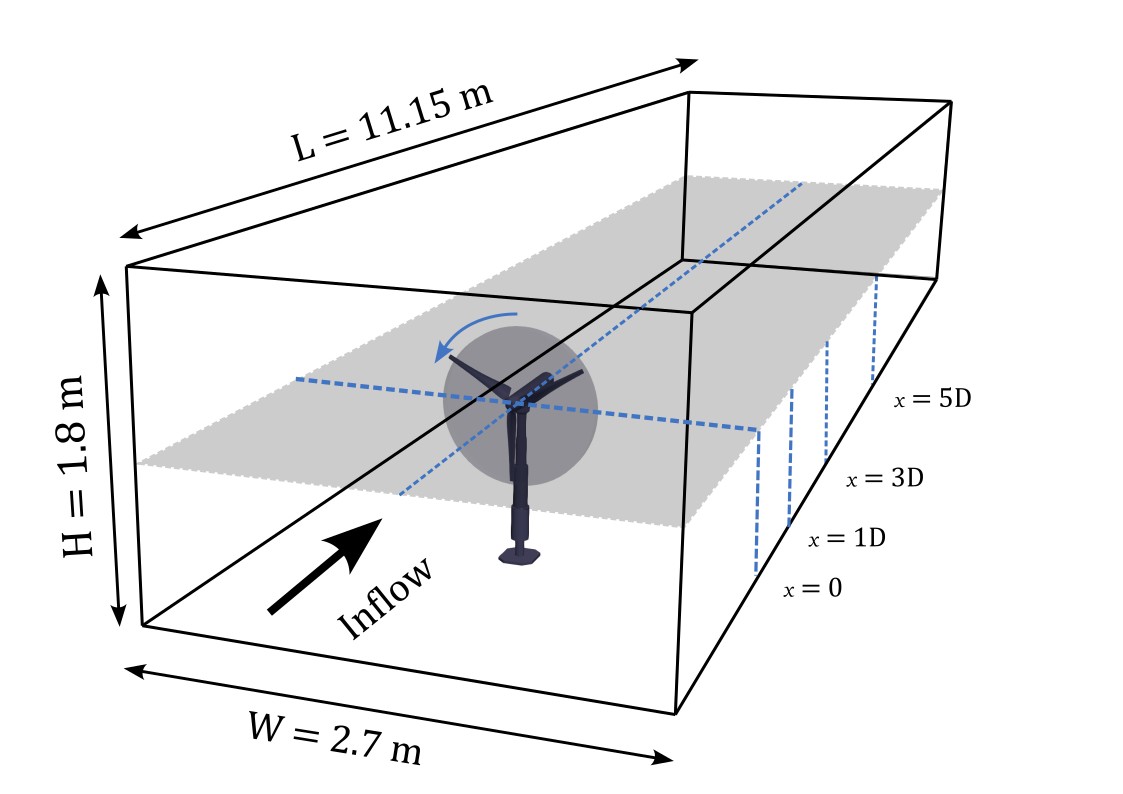}
    \label{fig:BT1_sche}}
    \subfloat[][Schematic representation of NTNU BT2.]{
    \includegraphics[width=0.45\textwidth]{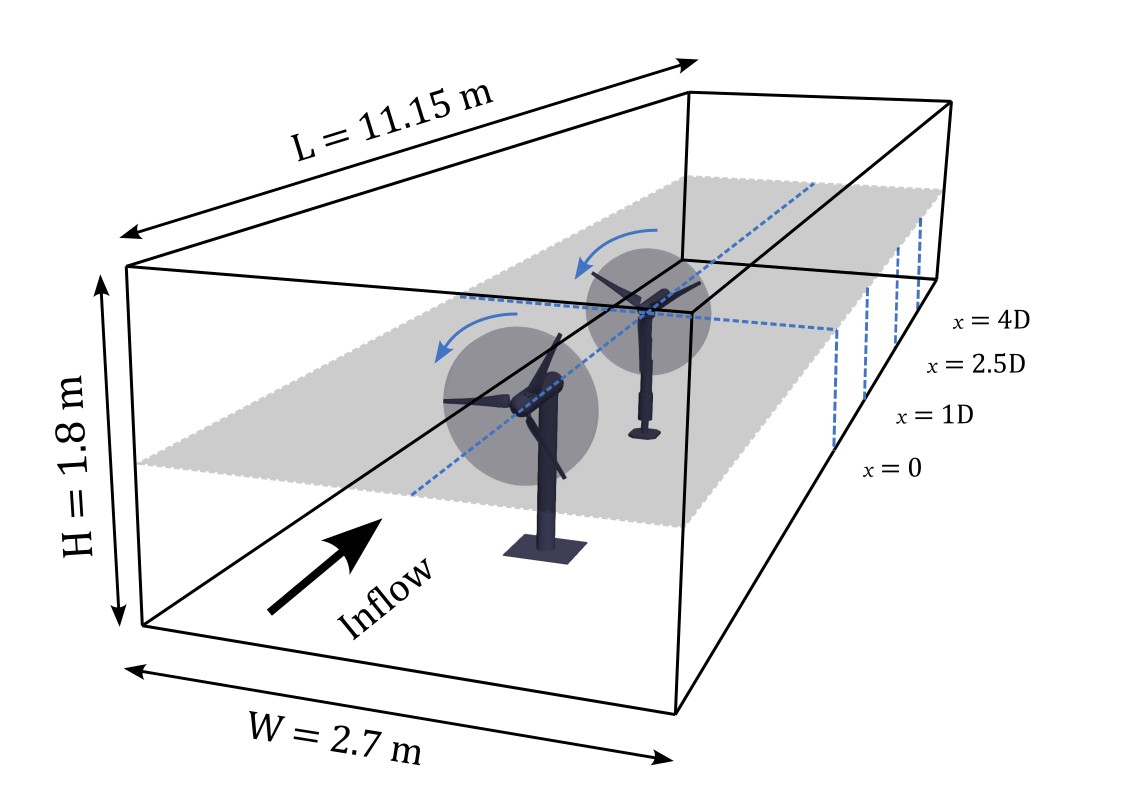}
    \label{fig:BT2_sche}}
    \qquad
    \subfloat[][Horizontal wake profile comparison -- BT1.]{
    \includegraphics[width=0.45\textwidth]{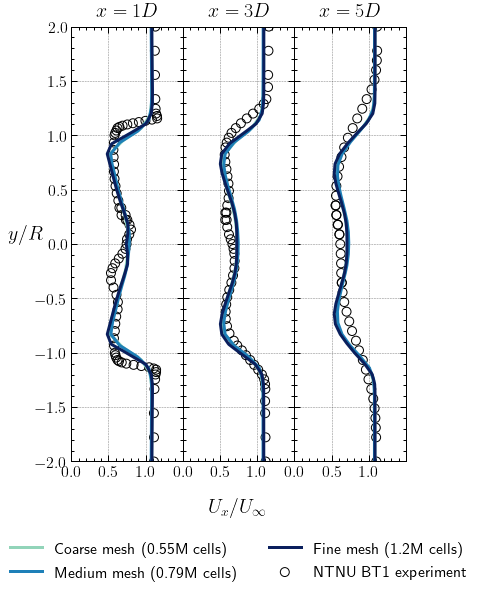}
    \label{fig:BT1_compare}}
    \subfloat[][Horizontal wake profile comparison -- BT2.]{
    \includegraphics[width=0.45\textwidth]{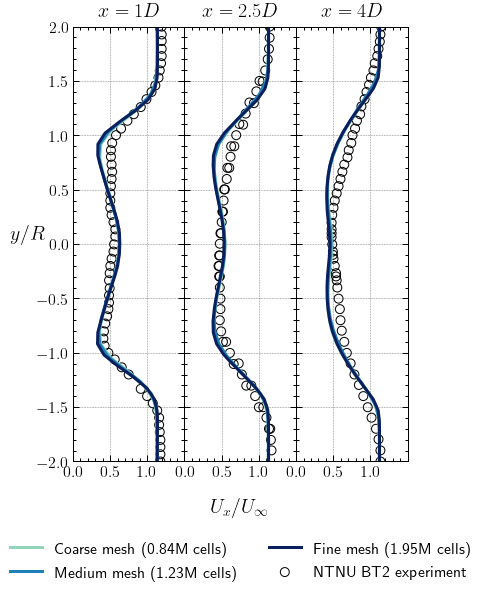}
    \label{fig:BT2_compare}}
    \caption{Schematic representation of the two setups of NTNU blind tests and the comparison of CFD simulation and experimental horizontal wake profiles.}
    \label{fig:blind_test}
\end{figure}

Both blind test experiments were conducted in a wind tunnel that was approximately 2.7 m wide, 1.8 m high and 11.15 m long. In BT1, a single model wind turbine with diameter $D = 0.894$ m was placed at a distance of 3.66 m from the wind tunnel inlet, whereas in BT2 two similar wind turbines with the same hub height but slightly different diameters ($D_1 = 0.894$ m and $D_2 = 0.944$ m, where the subscripts 1 and 2 correspond to the upstream and downstream turbines respectively) were mounted along the wind tunnel center line at a separation distance of $3D_1$, the upstream turbine was located at $2D_1$ from the wind tunnel inlet. Both experiments had the same operating conditions with a free stream wind speed of $U_{\infty} = \SI{10}{m.s^{-1}}$ and turbulence intensity of $\mathcal{I = } $ 0.3$\%$, and all wind turbines had three bladed rotors with the same blade geometry that used the NREL S826 airfoil. The turbine in BT1 had a design tip speed ratio (TSR) of 6, and wake statistics were measured at locations $x = $ 1, 3, and 5$D$ downstream of the turbine rotor. BT2 had the upstream turbine operating at $\textrm{TSR}_1$ = 6 and the downstream turbine at $\textrm{TSR}_2$ = 4, with wake statistics recorded at $x = $1, 2.5 and 4$D$ behind the downstream turbine rotor. Schematic representations of the two experiments are shown in Figures \ref{fig:BT1_sche} and \ref{fig:BT2_sche}. 

GAD simulation results with the realisable $k-\varepsilon$ turbulence model were compared against experimental data from NTNU, shown in Figures \ref{fig:BT1_compare} and \ref{fig:BT2_compare}. The choice of realisable $k-\varepsilon$ turbulence model was due to its better fit with experimental data compared to the standard $k-\varepsilon$ model, as is also reported in various other studies \cite{real_k_e_1, real_k_e_2}. The simulations were run with different mesh resolutions to check for mesh independence of the obtained solutions. An overall good agreement between GAD simulations and NTNU experimental data was found, which further demonstrated the ability of the GAD framework to model individual and multiple turbine interactions.  

\section{Graph Representation Learning}
\label{graph_representation_learning}
Many physical, biological and social systems can be described, at a certain level of abstraction, as graphs. In the case of CFD simulations, a finite computational mesh representing the spatial domain can be naturally defined as a graph, and as a consequence graph neural network (GNN) methods have great compatibility with CFD simulation data and can operate directly on unstructured meshes. Importantly, this eliminates the need for interpolation onto uniform grids, which is required for most CNN networks. This section details the graph neural network framework developed here for wake modelling, including training data generation, converting CFD data to appropriate graph data structures, the fundamentals of graph neural network theory and the proposed model architecture as well as relevant training experiments. The models were implemented using the open-source deep learning library PyTorch \cite{pytorch} and PyTorch Geometric \cite{pyg}, which is a geometric deep learning library based on PyTorch.

In order to be able to compare the geometric deep learning approach with a real world scenario, this work considered Sweden's Lillgrund offshore wind farm as a benchmark problem. Training data was generated based on CFD simulations under various operating conditions (inlet velocity, turbulence intensity and turbine yaw angle) of a stand-alone Siemens SWT93-2.3 MW wind turbine, which are deployed in the Lillgrund farm. CFD simulations were also performed on different rows of the Lillgrund wind farm to compare against superimposed single wake fields produced from the deep learning model.

\subsection{Data generation}
\begin{figure}[H]
    \centering
    \includegraphics[width=0.45\textwidth]{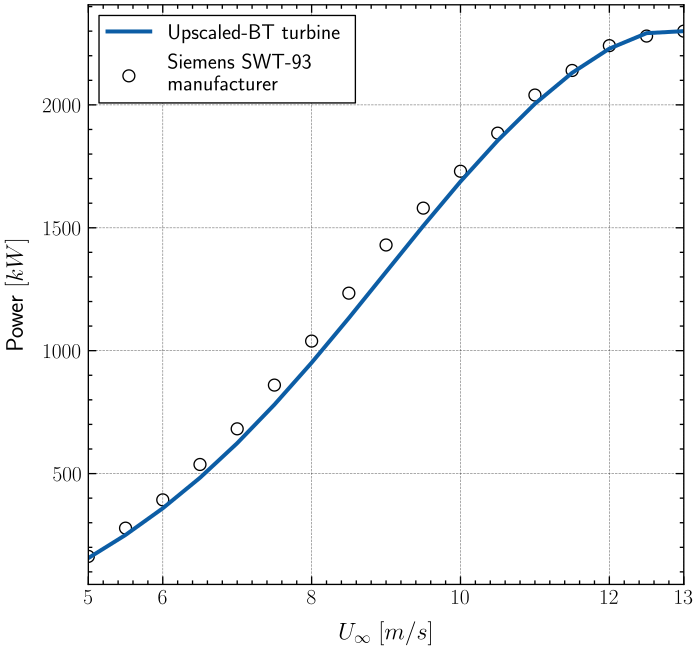}
    \caption{Power curve of the up-scaled ``blind test'' turbine compared to the manufacturer power curve of the Siemens SWT93-2.3 MW wind turbine.}
    \label{fig:power_curve}
\end{figure}

Training data was generated based on simulations of the Siemens SWT93-2.3 MW wind turbine, which had a turbine rotor with radius $R=46.5$m and hub height $H_{\text{hub}}=65$m. The exact blade geometry and airfoil characteristics have not, to the best of our knowledge, been disclosed to the public. As a consequence, an up-scaled version of the ``blind test'' turbine chord profile and airfoil characteristics were used, in a similar approach to \cite{yorgos}. This choice could be partially justified by the good agreement between the power curve for the up-scaled ``blind test'' turbine and the Siemens SWT93-2.3 MW turbine according to its manufacturer \cite{lillgrund_1}, as shown in Figure \ref{fig:power_curve}. 

\begin{figure}[H]
    \centering
    \includegraphics[width=\textwidth]{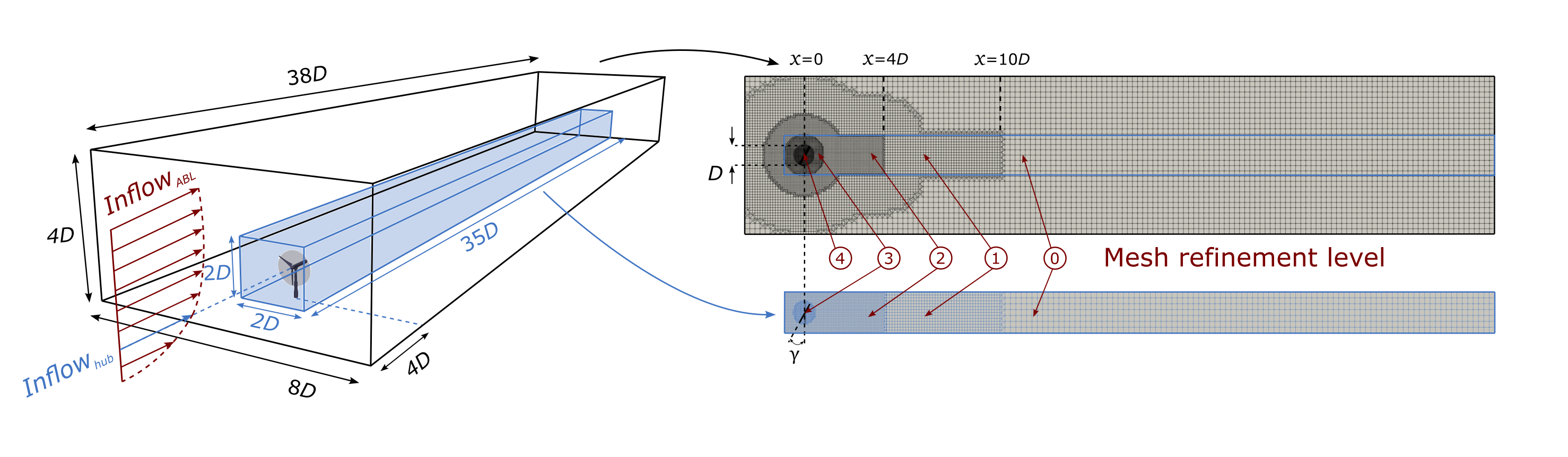}
    \caption{3D representation and 2D mesh of a horizontal slice (at hub height) of the computational domain for CFD simulations. Training data was extracted from the area shaded in blue. }
    \label{fig:training_data_domain}
\end{figure}

Training data was generated by varying the inflow velocity $U_{\infty, \;\text{hub}}$ and turbulent intensity $\mathcal{I}_{\infty, \;\text{hub}}$ at hub height as well as turbine yaw angle $\gamma$, the ABL boundary condition varied for each run with  $U_{\infty, \;\text{hub}}$ and $\text{TKE}_{\infty, \;\text{hub}} = \frac{3}{2} \mathcal{I}_{\infty, \;\text{hub}}^2 U_{\infty, \;\text{hub}}^2$ set as the reference velocity and TKE, and hub height set as reference height. The simulation domain was set to be sufficiently wide and high in order not to have an impact on simulation results, based upon sensitivity testing, and long enough to cover the longest row of the Lillgrund wind farm. For the deep learning task, training data was extracted from a smaller section of the simulation domain as shown in Figure \ref{fig:training_data_domain}. The motivation for this is that it would be sufficient for the deep learning model to learn directly from the reference domain that was embedded within a simulation of a larger domain, and that it is unnecessary for the deep learning model to be able make predictions on the entire simulation domain, but rather to concentrate on the area close to and behind the wind turbine where other turbines will likely be located.  The size of the training data domain was tested to be large enough to capture the entirety of the wake structure, including in the yawed cases. The computational mesh for both running CFD simulations and training deep learning models had several levels of resolution across the domain. As detailed in Figure \ref{fig:training_data_domain}, the mesh was densest in areas within and around the wind turbine actuator disk, and gradually coarsened at distance from these. A spherical refinement region with diameter 1$D$ was placed at the location of the wind turbine so that the same mesh could be used for simulations with different turbine yaw angles. The CFD simulation mesh had 0.6 million cells and a narrower area of 0.1 million cells was clipped from the simulation domain to be used as training data. Data from a total of 7700 simulations was generated, with the inflow velocity ranging from $5$ m/s to $10$ m/s, turbulence intensity from $5\%$ to $15\%$ and yaw angle from $-30^{\circ}$ to $30^{\circ}$. The 7700 simulation data
took about 4000 CPU hours to generate on Imperial College London's CX1 HPC cluster, with each simulation taking on average approximately 0.5 hours for the SimpleFoam steady-state solver to converge, and was partitioned into sets of size 6200/750/750 for training, validation and testing respectively.

\subsection{Graph neural networks}
This work utilised graph neural networks for a supervised graph representation learning task, and took as input a graph $G = (V, E)$, where $V \in \mathbb{R}^{n \times f}$ represents the $n$ vertices with $f$ features on each vertex, and $E \in \mathbb{R}^{2 \times n_{e}}$ stands for the $n_{e}$ number of edges, represented via the pairs of vertices that form each of them. The computational mesh directly defines $E$ and can be converted to graph structure by appending the $(x, y, z)$ coordinates of the mesh vertices as well as their corresponding one-hot encoded boundary types as vertex features in $V$. Additionally the physical parameters that define the flow structure and vary among each simulation (i.e. inlet velocity $U_{\infty}$, turbulence intensity $\mathcal{I}_{\infty}$ and turbine yaw angle $\gamma$) are also appended to each vertex in $V$ as global features, so that the network can differentiate among simulation training data and learn to make predictions based on various input physical parameters. Target flow fields were associated with each vertex as output responses. 

\begin{figure}[H]
    \centering
    \includegraphics[width=0.65\textwidth]{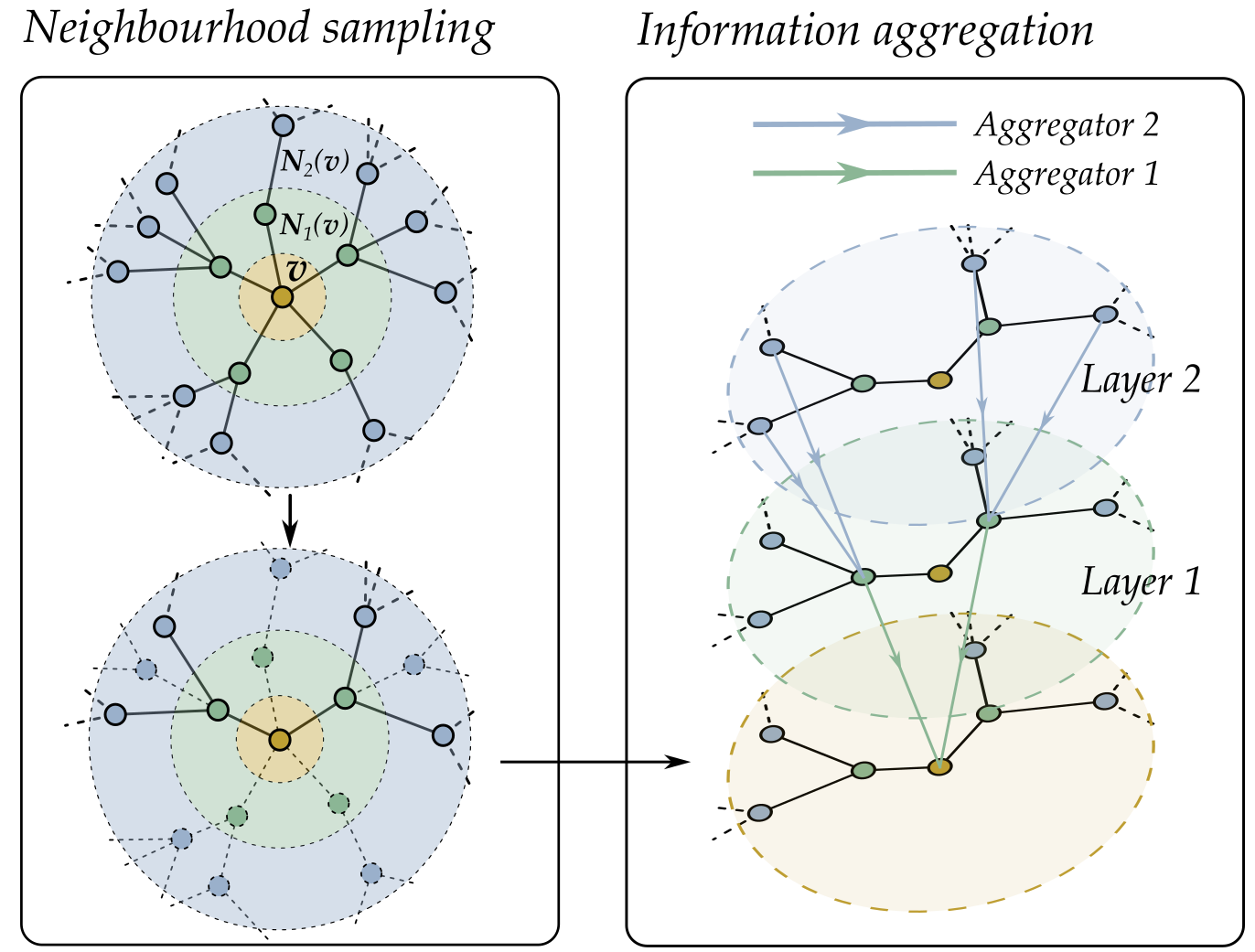}
    \caption{Visual illustration of the GraphSAGE sample and aggregation approach in a two-layer case for a target vertex $v$. $N_1(v)$ and $N_2(v)$ represent the one-hop and two-hop neighbourhoods of $v$, respectively.}
    \label{fig:sage_illus}
\end{figure}

This work made extensive use of the GraphSAGE framework due to its strong inductive learning capabilities and ability to scale very well on larger graphs \cite{graphsage_paper, graphsage_book}. 
The GraphSAGE neural network with mean aggregation was used, the vertex-wise update rule of the $k$-th layer of the GraphSAGE network with mean aggregation on a vertex embedding $\mathbf{x}_v$ can be written as:
\begin{fleqn}
    \begin{equation}
        \begin{aligned}
         &\mathbf{x}^{(k)}_{v} = \sigma\left(\mathbf{W}^{(k)} \left[ \mathbf{x}_v^{(k-1)}
        \oplus\mathbf{x}_{\mathcal{N}(v)}^{(k)} \right]\right),\\
        &\mathbf{x}_{\mathcal{N}(v)}^{(k)} =   \frac{1}{|{\mathcal{N}(v)}|} \sum_{u \in  \mathcal{N}(v)} \mathbf{x}_u^{(k-1)},
        \end{aligned}
    \end{equation}
\end{fleqn}    
where $\mathcal{N}(v)$ is a fixed-sized sampled neighbourhood of vertex $v$ used to aggregate information, and not the full neighbourhood, $\sigma(\cdot)$ is a non-linear activation function, $\mathbf{W}^{(k)}$ is the weight matrix of $k$-th layer, and $\oplus$ stands for concatenation. An illustration of the GraphSAGE network is shown in Figure \ref{fig:sage_illus}.

Various other modern GNN architectures, most notably the standard graph convolutional network (GCN) \cite{gcn} and graph attention network (GAT) \cite{gat}, were also explored in this work as potential alternatives to GraphSAGE. The GCN update rule can be written as:

\begin{fleqn}
    \begin{equation}
       \mathbf{x}^{(k)}_v = \sigma\left(\sum_{u \in \mathcal{N}(v)}\frac{1}{\sqrt{|\mathcal{N}(u)|\cdot|\mathcal{N}(v)|}}\mathbf{W}^{(k)} \mathbf{x}_u^{(k-1)}\right),
    \end{equation}
\end{fleqn}    
whereas the update rule for GAT is:
\begin{fleqn}
    \begin{equation}
         \mathbf{x}^{(k)}_{v} = \oplus_{h=1}^H \sigma\left(\sum_{u \in \mathcal{N}(v)}\alpha_{uv}^{(h)} \mathbf{W}^{(h)} \mathbf{x}^{(k-1)}\right),
    \end{equation}
\end{fleqn}  
where $H$ is the number of attention heads and $\alpha_{uv}$ represents the attention coefficient. GAT is a powerful model that uses the attention mechanism to allow for the implicit assignment of different levels of importance to a vertex’s neighbours, therefore leading to a sizeable increase in model capacity \cite{gat}. It should be noted that while GraphSAGE performs layer-wise sampling from the neighbourhood of each vertex, other GNN methods including GCN and GAT, use the full neighbourhood of vertices. 

One limitation of deep graph neural networks is over-smoothing, which refers to similarity of vertex representations after several iterations of message passing. This can occur when more layers are added to the structure, as shown in Figure \ref{fig:sage_illus}, and eventually every vertex in the graph is able to aggregate information from distant neighbours therefore generating similar graph embeddings. Indeed, various modern GNN models including GCN and GAT achieved their best performance on benchmark problems with models that had only two layers. In particular, CFD data on meshes could have drastically different resolutions across the simulation domain, as a consequence vertices might also need to aggregate information from neighbours of different distances depending on their local spatial mesh resolution. The reason being that vertices in an area of lower mesh density might be closer to each other in graph space but very far apart in mesh space, and similarly vertices in regions with denser meshes might be far from each other in graph space, but very close in mesh space. In order to alleviate over-smoothing, this work adopted jumping knowledge connections (JK) \cite{JKNet, GCN_res} which uses dense skip-connections to enable the adaptive learning of structure-aware vertex representations, as well as layer-wise residual connections similar to ResNet \cite{resnet} and an additional initial residual connection inspired from GCNII \cite{GCNII}.

\begin{figure}[H]
    \centering
    \includegraphics[width=\textwidth]{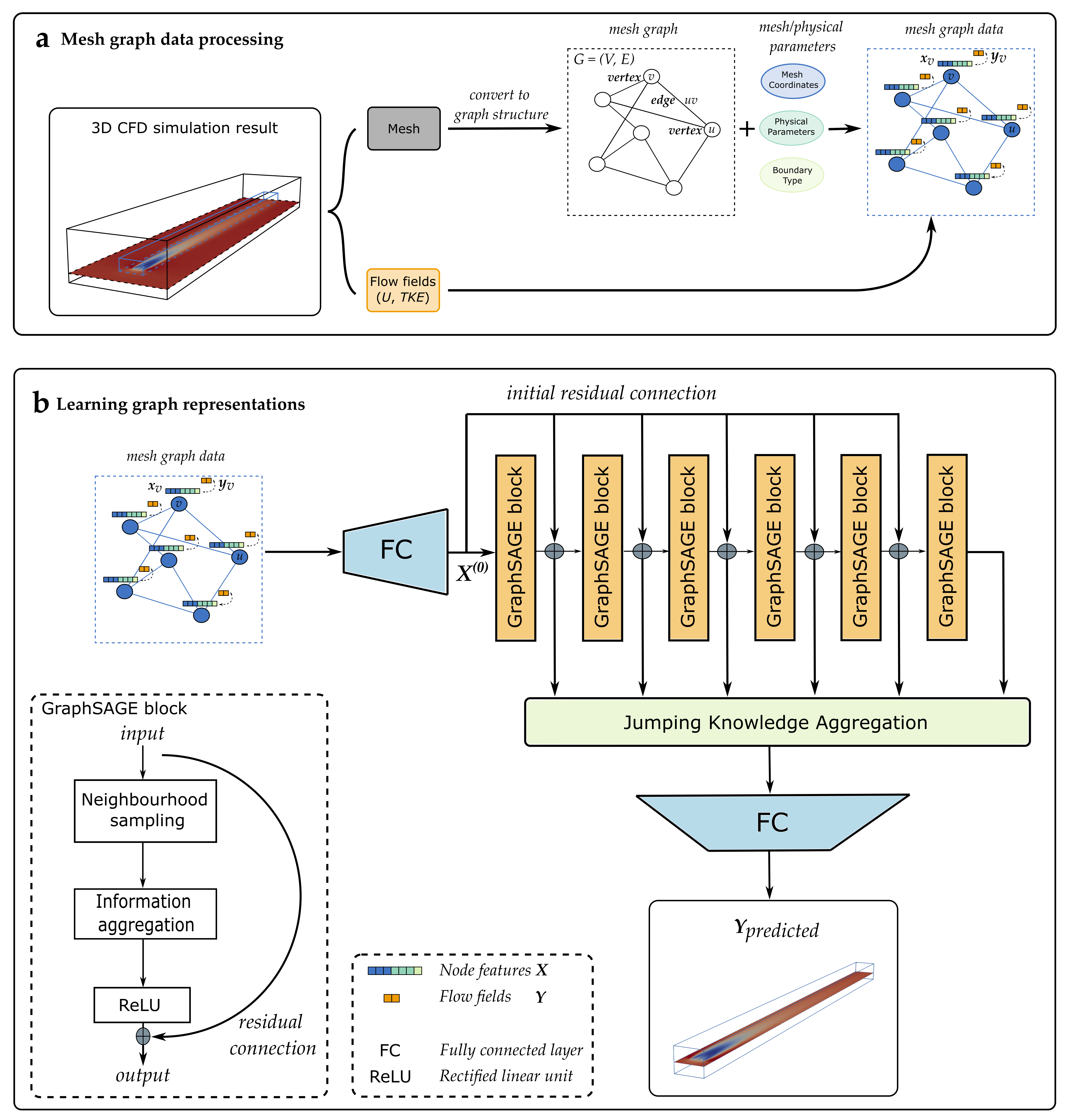}
    \caption{The overall workflow of converting CFD simulation data to mesh graph data and the GraphSAGE graph neural network architecture with jumping knowledge and residual connections.}
    \label{fig:WakeML_workflow}
\end{figure}

Together with residual connections, the vertex-wise update rule of the weighted GraphSAGE network with mean aggregation can be written as:

\begin{fleqn}
    \begin{equation}
      \mathbf{x}^{(k)}_v = \sigma\left(\mathbf{W}^{(k)} \left[ \mathbf{x}_v^{(k-1)} 
      \oplus \frac{1}{|{\mathcal{N}(v)}|} \sum_{u \in \mathcal{N}(v)}  \mathbf{x}_u^{(k-1)} \right]\right) + \alpha  \mathbf{x}^{(0)}_v + \beta  \mathbf{x}^{(k-1)}_v,
    \end{equation}
\end{fleqn}    
where $\alpha$ and $\beta$ are hyper-parameters that signify scales of residual representations from previous layers, and were set to 0.1 and 0.9 respectively in this work. The network architecture of GraphSAGE with jumping knowledge  and residual connections is illustrated in Figure \ref{fig:WakeML_workflow}b.

\subsection{Supervised training experiments}
\begin{figure}[H]
    \centering
    \includegraphics[width=\textwidth]{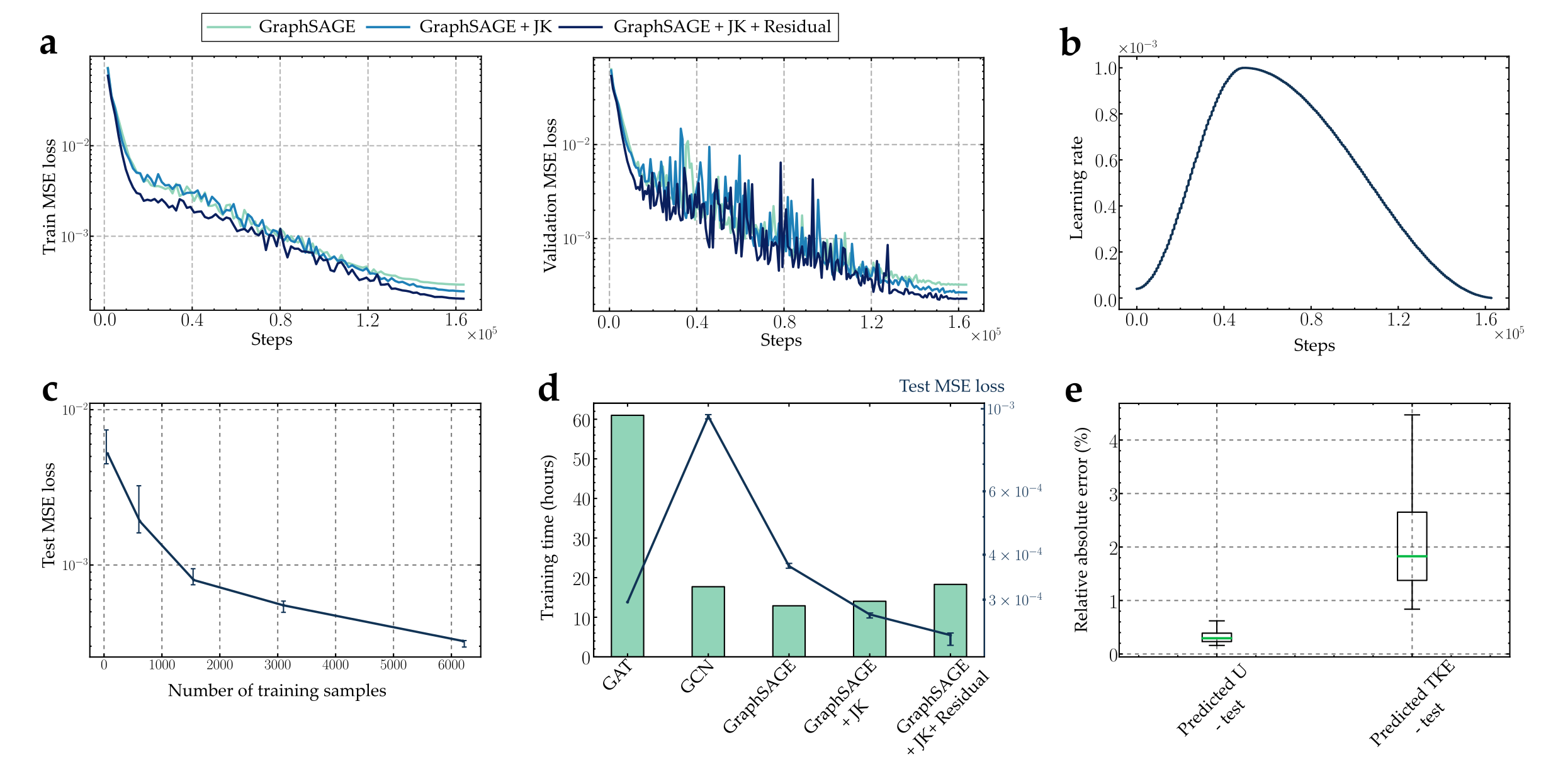}
    \caption{Supervised training experiments on GNN architectures. $\mathbf{a}$: Training and validation learning curves for the three GraphSAGE variants. $\mathbf{b}$: The one-cycle learning rate scheme -- learning rate was set to start at a small value and climbs up to reach a pre-defined maximum value before gradually decreasing. $\mathbf{c}$: Training of the standard GraphSAGE model with different number of training samples. $\mathbf{d}$: Training time and MSE loss on the unseen test set for different graph neural network models. Training time and test MSE loss for GAT were taken from a single run due to the excessive amount of time needed for training, while for other models training time was averaged over three runs and test MSE loss was reported with error bars. $\mathbf{e}$: Box-and-whisker plots of relative absolute error of the final model at predicting $U$ and TKE flow fields on the held-out test data set. The whiskers represent 1.5 times the inter-quartile range (IQR). Green line represents median accuracy on test data.}
    \label{fig:experiments}
\end{figure}

Training experiments and ablation studies were performed in an attempt to improve model performance. Unless otherwise specified, all experiments performed in this work were trained to minimise the mean squared error (MSE) of the predicted flow fields, with the AdamW optimiser \cite{adamw} with the one-cycle \cite{one_cycle} learning rate scheduler for 160k steps. With the maximum learning rate set to $10^{-3}$ the one-cycle learning rate schedule yielded very good training convergence with all models and ensured that different training runs were comparable; the learning rate scheme is illustrated in Figure \ref{fig:experiments}b. A batch size of one was used due to limited GPU memory, gradients were accumulated over 16 mini-batches in compensation. 16 bit automatic mixed precision (AMP) training was used to speed up training and save GPU memory, validation error was checked twice per epoch and models were checkpointed each time upon reaching a lower validation error. All experiments were run on a single NVIDIA GeForce RTX 2080 GPU with 8 GB of memory three times with different random seeds, and when comparing different models, all models comprised of six GNN layers with 128 hidden units in each layer, with a total of trainable 220k parameters.

The core results from the experiments are shown in Figure \ref{fig:experiments}. Figure \ref{fig:experiments}a shows an ablation study of the three variants of the GraphSAGE model, and demonstrates that jumping knowledge and residual connections led to improved model performance, as the GraphSAGE model managed to reach lower training and validation loss after adding a JK aggregation layer and residual connections. The amount of data required for model training was investigated and the relevant findings are reported in Figure \ref{fig:experiments}c. The amount of training data utilised was varied from 60 samples to 6200 samples, and the performance of the GraphSAGE model was reported after training with samples of different sizes for a fixed number of 160k steps. It can be observed that the graph neural network is able to achieve relatively low MSE error even when trained with a small fraction of the full training data, with the test loss continuing to decrease when more data is used in the training process. The effect of number of training samples on model performance is further illustrated in Figure \ref{fig:num_data_comparisons}, where flow field predictions from models trained with different numbers of training samples were compared for a given unseen test case. Noticeably even with a relatively small number of training samples the standard GraphSAGE model is able to correctly capture the majority of the primary features of the fluid flow. The near-wake region has the densest mesh and has the most impact on the model performance criterion, and machine learning predictions (particularly around the near-wake area) became more accurate as more training samples were included. It is also manifest that the utilisation of jumping knowledge and residual connections led to markedly better predictions around regions of changing mesh resolution. It should be noted that while this work kept the mesh in the near wake regions dense in order to preserve the entirety of the fluid flow structures, it is possible to customise the mesh used during training, and to further refine or coarsen the mesh in areas that are of more or less significance. An exploration of fully exploiting the combined flexibility of both unstructured mesh CFD and GNNs will be explored in future work.

\begin{figure}[H]
    \centering
    \subfloat[h][Horizontal slice of predicted velocity magnitude and the corresponding error map when different number of training samples were used.]{
    \includegraphics[width=0.95\textwidth]{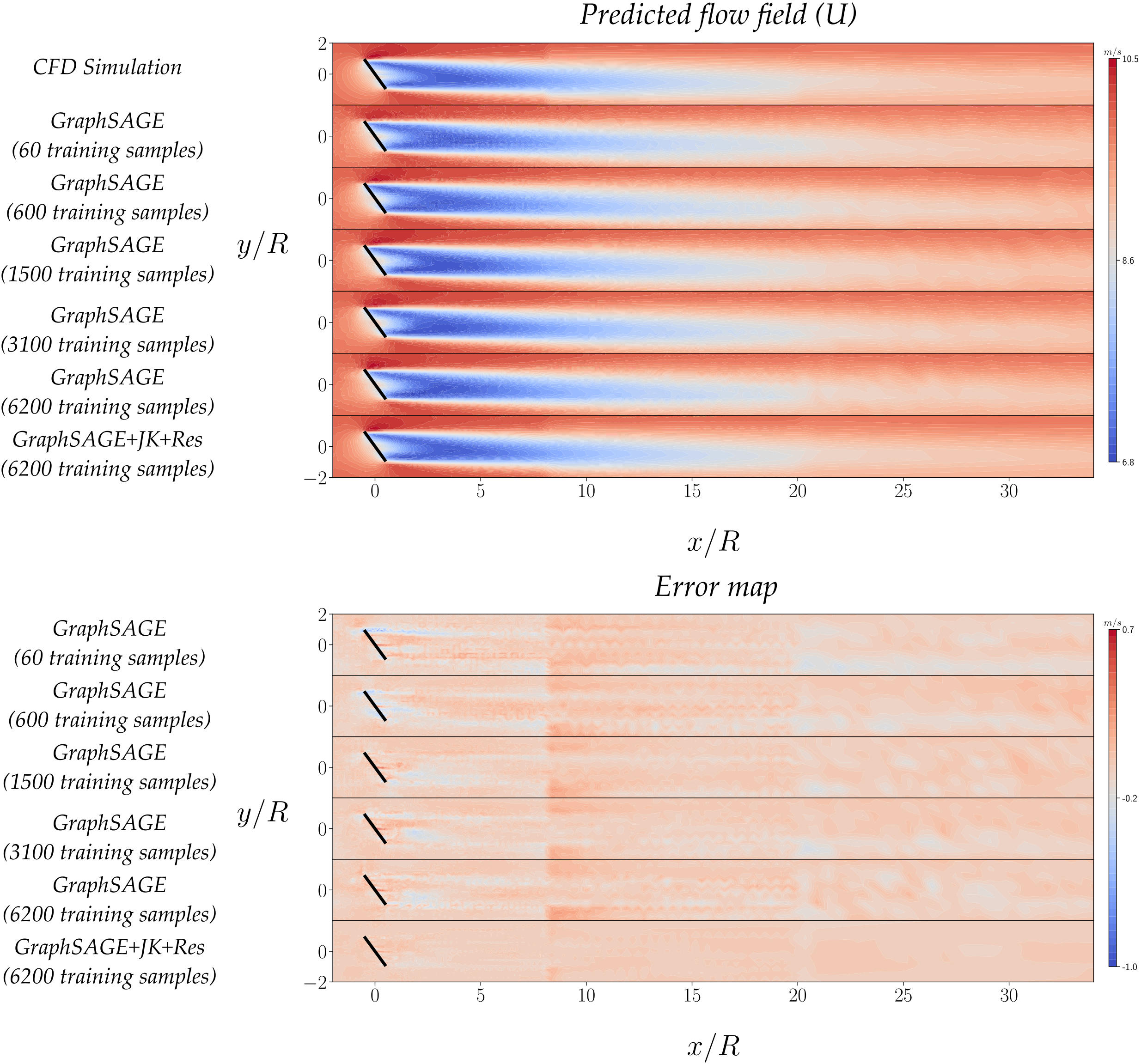}
    \label{fig:num_data_U}}
    \setbox0=\vbox{\caption{}}
\end{figure}

\begin{figure}[]
    \ContinuedFloat
    \subfloat[h][Horizontal slice of predicted TKE and the corresponding error map when different number of training samples were used.]{
    \includegraphics[width=0.95\textwidth]{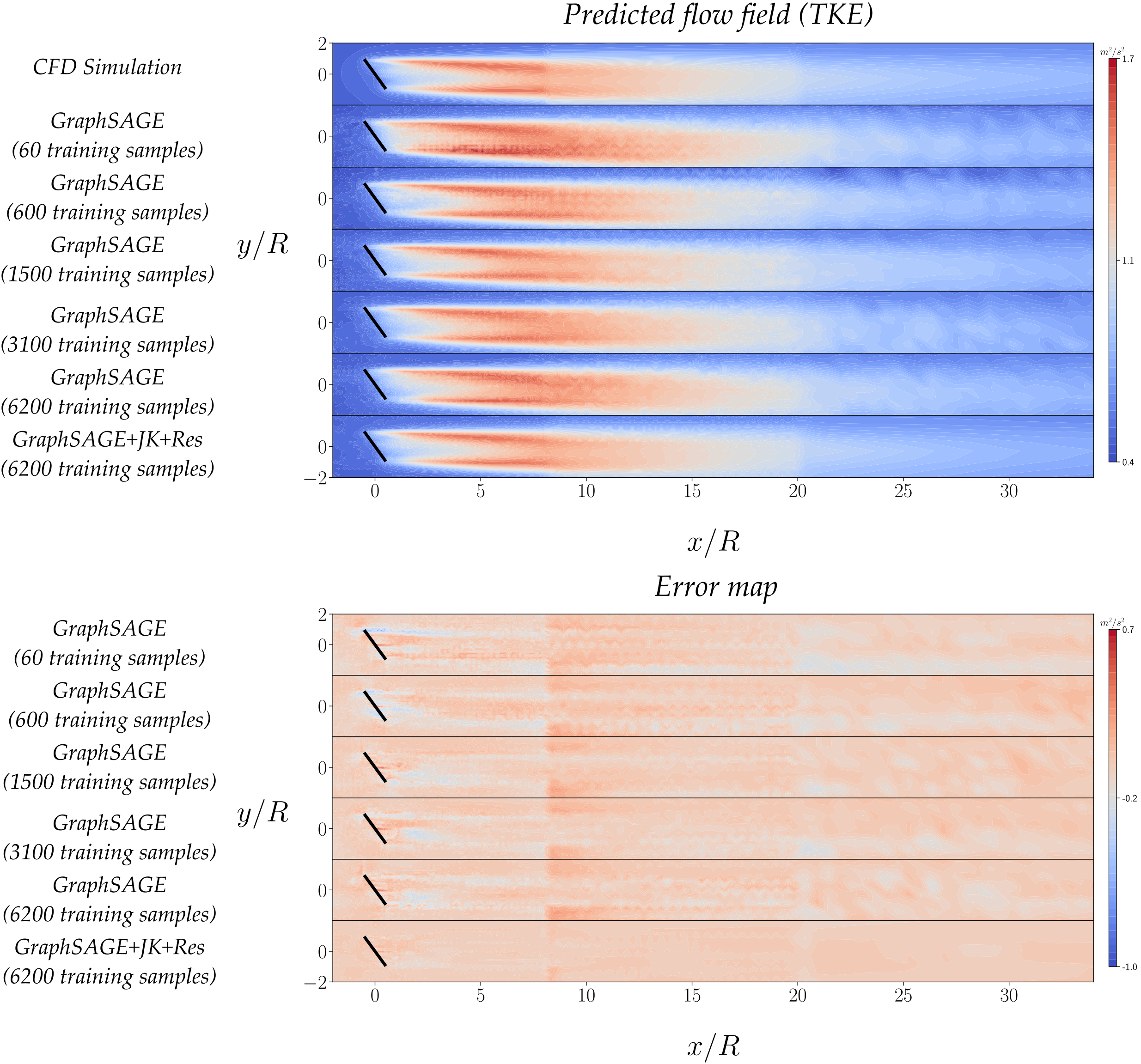}
    \label{fig:num_data_TKE}}
    \caption{Illustration of the effect of number of training samples on predictions by the GraphSAGE model and a comparison between GraphSAGE and GraphSAGE+JK+Res model predictions. The case parameters used were $U_{\infty}$=10 m/s, $I_{\infty}$=7.1\%, $\gamma$=$-28^{\circ}$.}
    \label{fig:num_data_comparisons}
 \end{figure}

All GraphSAGE variants proved to scale very well with the large data sizes used in this work, with the GraphSAGE+JK+Residual model taking approximately the same time to train as the standard GCN model, while achieving a much lower test loss. In contrast the GAT model, despite managing a lower test loss than the standard GCN and GraphSAGE model, took considerably longer time to train, as shown in Figure \ref{fig:experiments}d. The GraphSAGE+JK+Residual model was able to attain better accuracy than GAT, while taking only $1/3$ of the training time. It is worth noting that jumping knowledge and residual connection could also be used with GAT to further improve performance, but the training time is prohibitive, and the increase in training time could be further exacerbated when larger data sizes are considered. The scalability advantages of the GraphSAGE variants are expected to become even more prominent as the size and scale of training data grow. Figure \ref{fig:experiments}e shows the performance of the GraphSAGE+JK+Residual model on the unseen test dataset. The model was trained more extensively for 320k steps, and achieved a median relative absolute accuracy of 99.71$\%$ in predicting $U$ and 98.17$\%$ in predicting TKE. The model was able to generalise well to unseen data and the vast majority of the predictions on the test dataset had relative absolute error of less than $5\%$. This model was considered the final trained model and several examples of its predictions are shown in Figure \ref{fig:cfd_gnn_comparisons}.  The final model had 220k parameters and can make predictions in 13.2 $ms$ $\pm$ 120 $\mu s$ on a single GPU (including the time needed to transfer data from the CPU to GPU) and 2.93 $s$ $\pm$ 207 $ms$ on an Intel Core i9-9900K 8 core CPU.

\begin{figure}[H]
    \centering
    \subfloat[h][Case 1 parameters: $U_{\infty}$=9.4 m/s, $I_{\infty}$=13.9\%, $\gamma$=$-16^{\circ}$, horizontal slice shown.]{
    \includegraphics[width=0.9\textwidth]{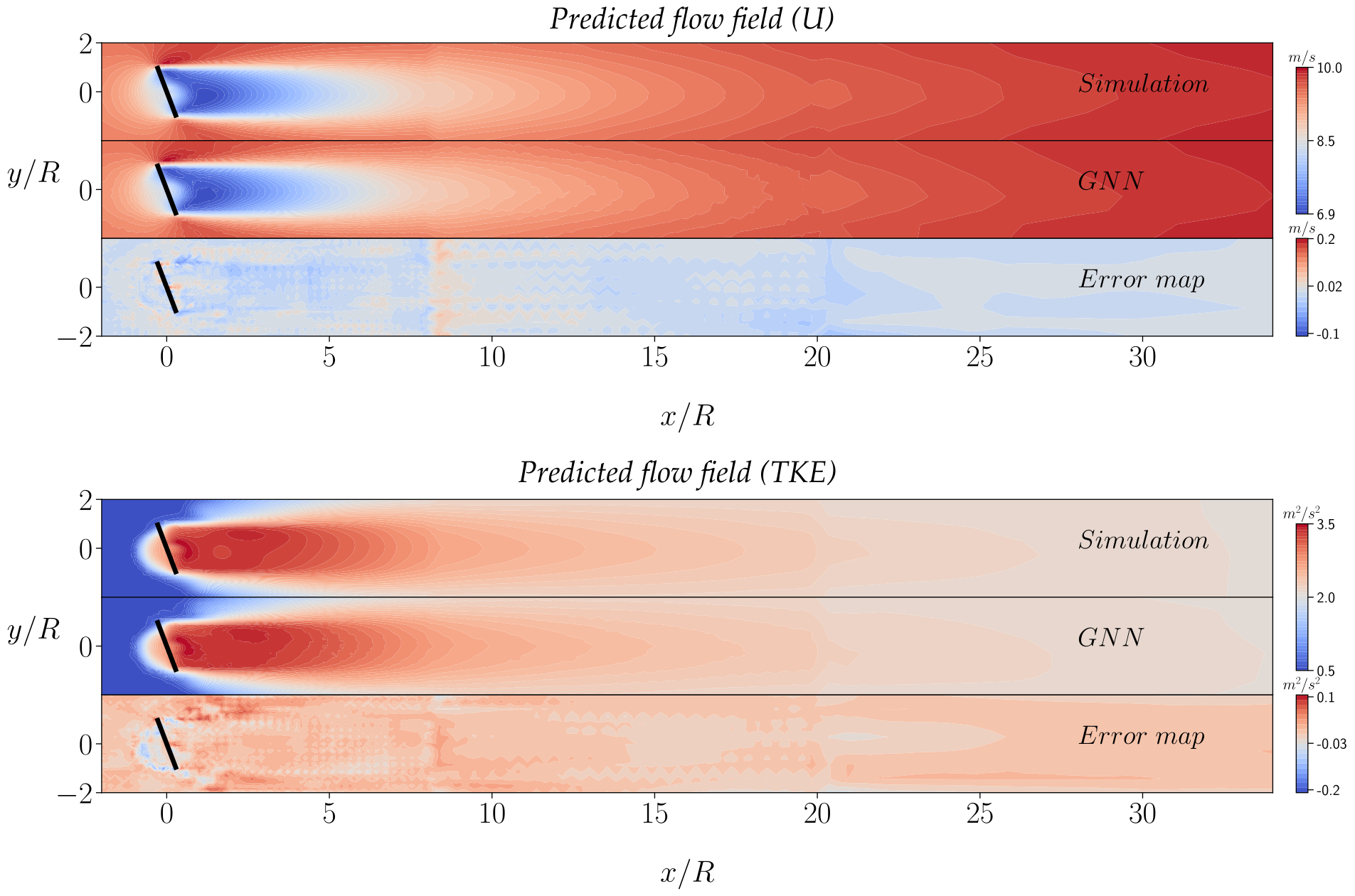}
    \label{fig:comparison_1}}
    \setbox0=\vbox{\caption{}}
\end{figure}
    
\begin{figure}[H]
    \ContinuedFloat
    \centering 
    \subfloat[h][Case 2 parameters: $U_{\infty}$=8.3 m/s, $I_{\infty}$=7.6\%, $\gamma$=$28^{\circ}$, horizontal slice shown.]{
    \includegraphics[width=0.9\textwidth]{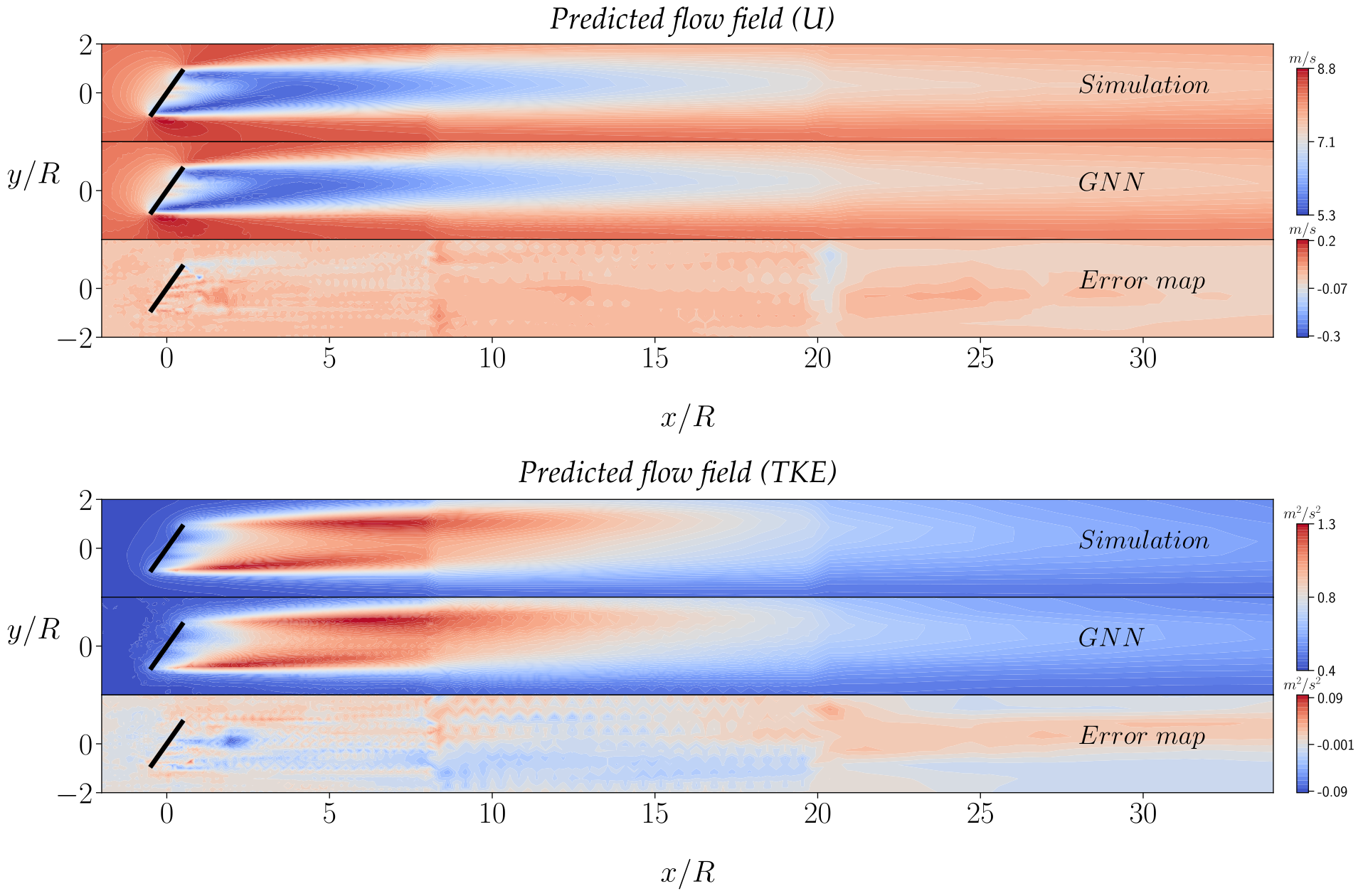}
    \label{fig:comparison_2}}
    \setbox0=\vbox{\caption{}}
 \end{figure}
 \begin{figure}[H]
    \ContinuedFloat
    \centering
    \subfloat[h][Case 3 parameters: $U_{\infty}$=9.4 m/s, $I_{\infty}$=9.2\%, $\gamma$=$-4^{\circ}$, vertical slice shown.]{
    \includegraphics[width=0.9\textwidth]{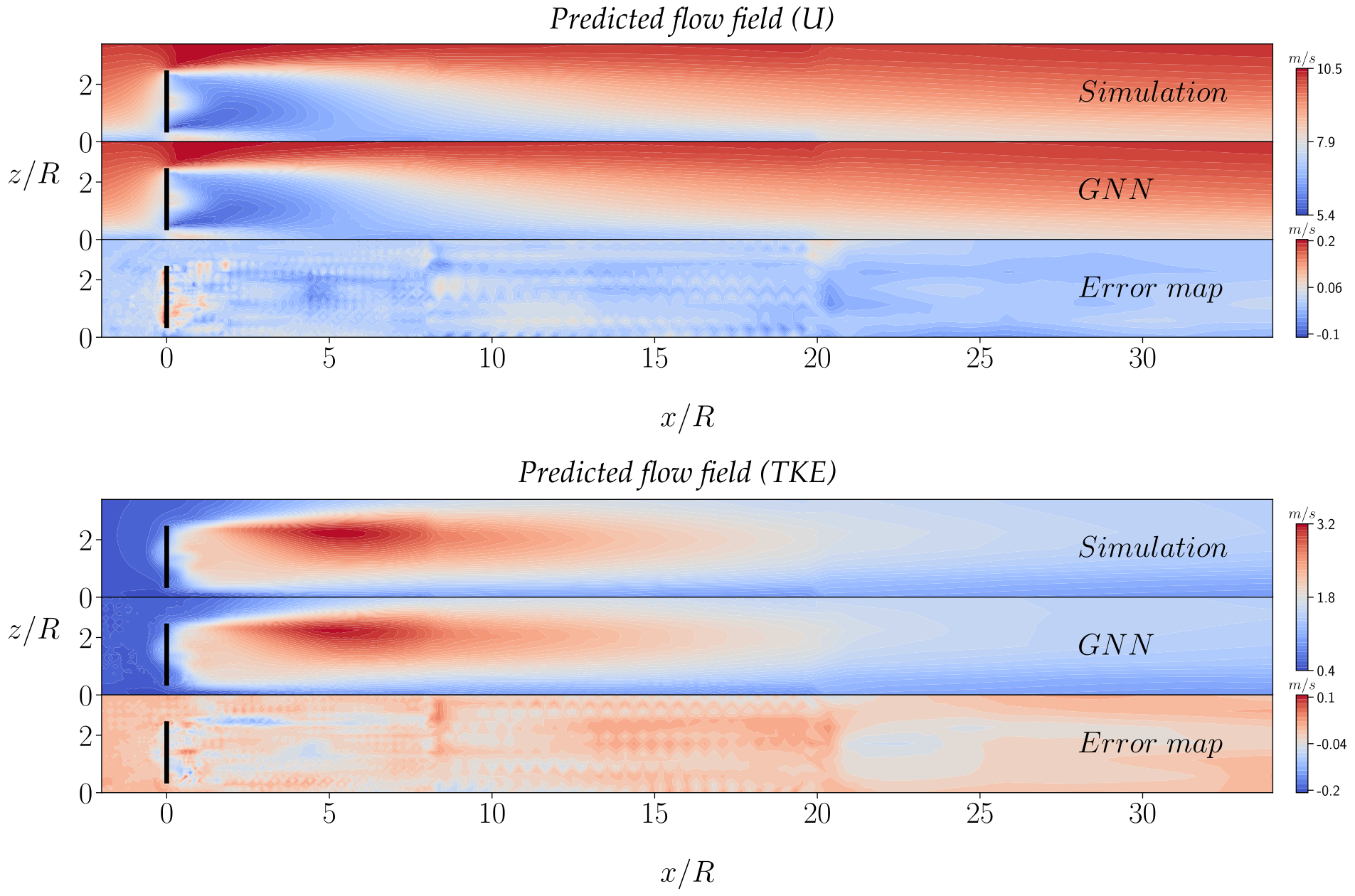}
    \label{fig:comparison_3}}
    \qquad
    \caption{Comparison between CFD simulation result and corresponding machine learning predictions made by the GraphSAGE+JK+Residual model illustrated in the form of 2D horizontal and vertical slices. Machine learning predictions were made on simulation parameters that were unseen during training. The top rows show the flow fields (velocity magnitude $U$ and TKE) computed from CFD simulation, the middle rows show the predictions made by the trained GNN model and the bottom rows display the difference between the CFD and machine learning predictions.}
    \label{fig:cfd_gnn_comparisons}
\end{figure}

\subsection{Model testing on multiple turbines}

The ability of GNN based wake model to model a real wind farm was tested by using the trained GNN model to predict individual turbine wakes, and superimposing the predicted wakes with a standard wake superposition approach. Specifically, the sum of squares superposition method was used, which has the following form:
\begin{fleqn}
    \begin{equation}
        U_i = \left({1 - \sqrt{\sum_{j=1}^{n_i}(1 - \frac{U_{ij}}{U_{\infty, j}}})^2}\right) U_{\infty},
    \end{equation}
\end{fleqn}
where $U_{i}$ is wind velocity at turbine $i$, $U_{ij}$ refers to the wind velocity at wind turbine $i$ influenced by the wake of wind turbine $j$, $U_{\infty,j}$ is the inlet velocity experienced by turbine $j$ and $n_i$ is the total number of upstream wind turbines. The choice of the optimal superposition method is dependent on the relative positions of the turbines \cite{superposition}. Other wake superposition methods proved unsuitable for this test case, with the linear and the largest deficit superposition methods underestimating and overestimating the wind velocity respectively.

\begin{figure}[H]
    \centering
    \includegraphics[width=0.8\textwidth]{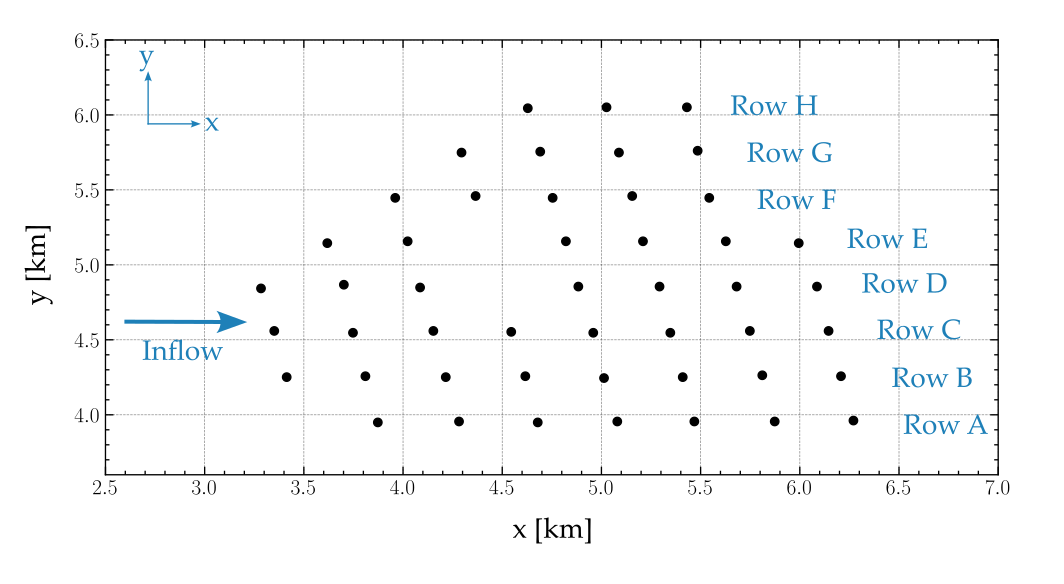}
    \caption{Lillgrund wind farm layout where the $x$-$y$ coordinate system were aligned with the southwestern statistically dominant wind direction.}
    \label{fig:layout}
\end{figure}

The wind farm investigated was the Lillgrund offshore wind farm, which consisted of 48 Siemens SWT93-2.3 MW wind turbines distributed in eight rows (A--H). CFD simulations were performed on different rows of the Lillgrund wind farm considering the statistically dominant wind direction and assuming independence of the rows. The resulting layout of the wind farm is shown in Figure \ref{fig:layout}. 

\begin{figure}[H]
    \centering
    \includegraphics[width=0.8\textwidth]{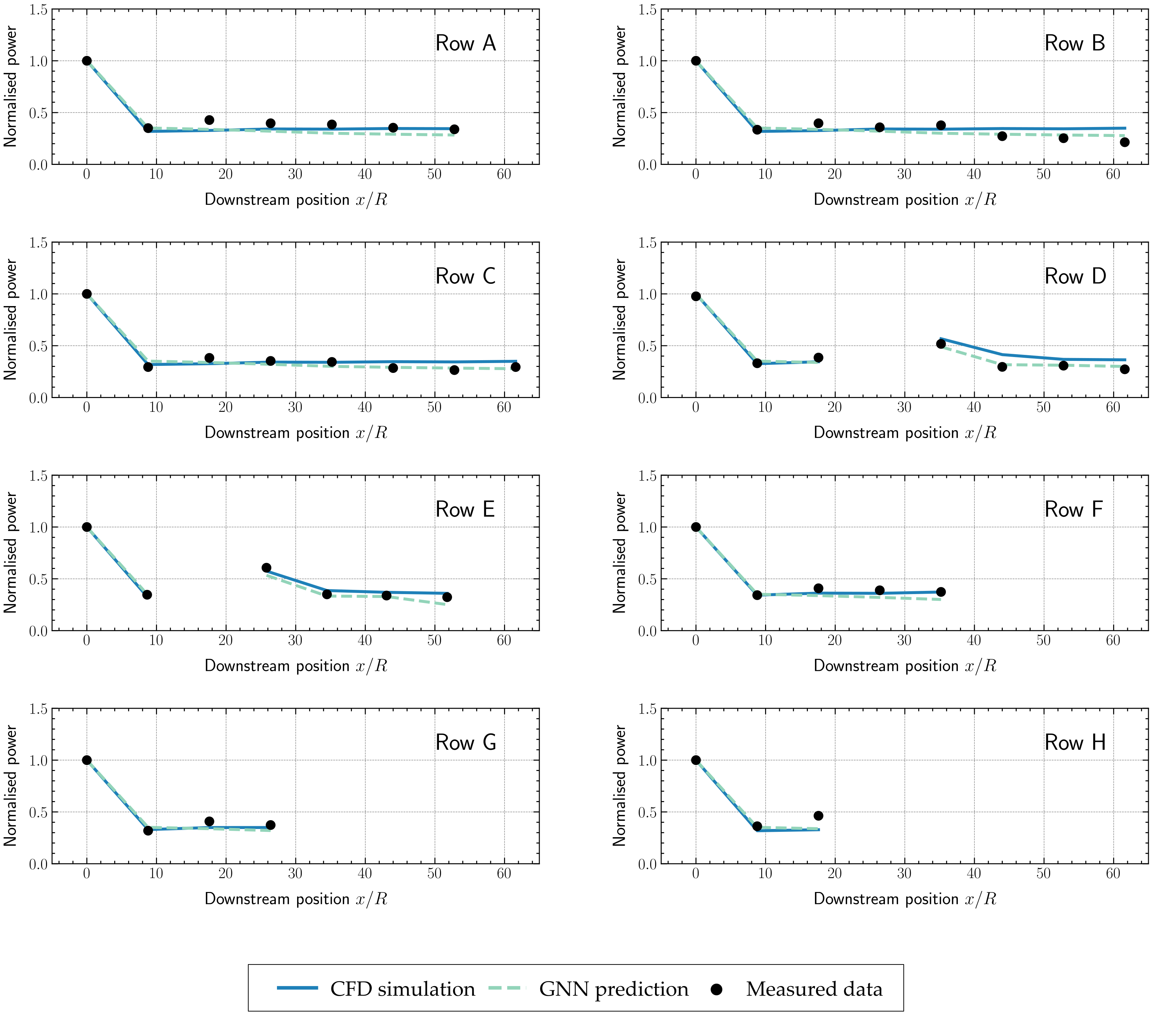}
    \caption{Comparison of predicted power from CFD simulations and GNN predictions with wake superposition against measured data for each row of the Lillgrund wind farm.}
    \label{fig:power_comparison}
\end{figure}

\begin{figure}[H]
    \centering
    \includegraphics[width=\textwidth]{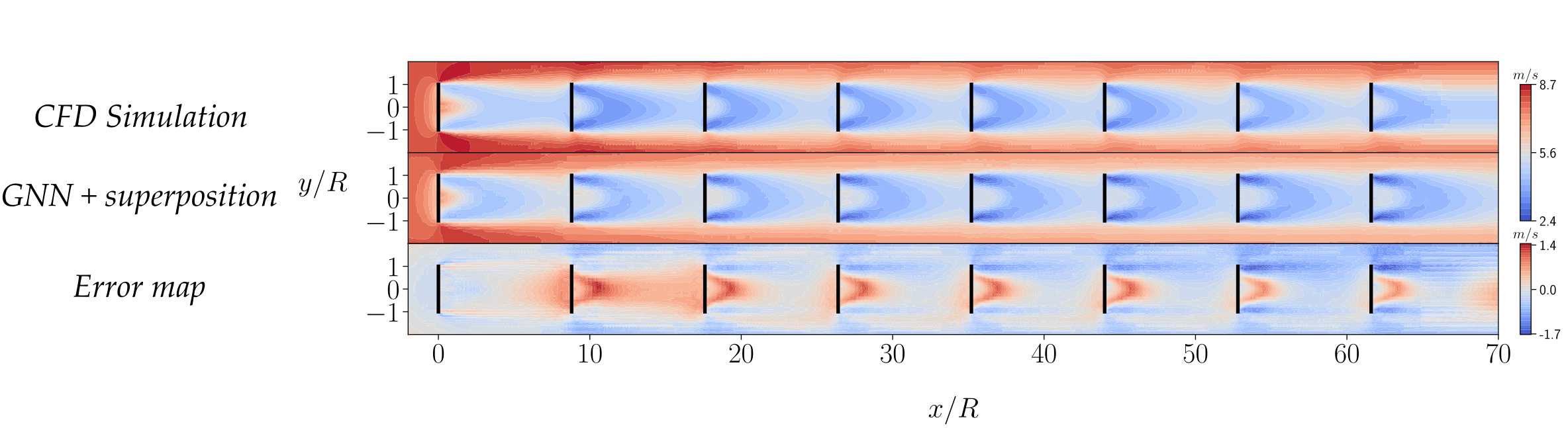}
    \caption{Velocity flow field of Row B of the Lillgrund wind farm. The top row shows the flow fields computed from a CFD simulation, the middle row shows the SOS superimposed predictions made by the GNN model and the bottom row displays the difference between CFD and machine learning predictions. }
    \label{fig:superposition_field}
\end{figure}

The power generated by each turbine in the GNN based wake model was computed as $P = \frac{1}{2}\rho C_p A U^3$, with $C_p$ given by the power curve shown in Figure \ref{fig:power_curve}. Power generated by wind turbines in different rows of Lillgrund from CFD simulations, the GNN based model and measured data \cite{lillgrund_1} are compared in Figure \ref{fig:power_comparison}. It can be observed that in general there is good agreement among all three of the CFD computations, GNN predictions and the measured data. In particular, the GNN wake model also managed to accurately capture the predicted power in rows D and E where there was a gap within the turbine rows. Nevertheless, superimposing individual turbine wakes led to a small underestimation in the generated power of turbines located far downstream compared to CFD, as can be observed in most rows of this test case. The velocity flow field predictions from CFD simulation and GNN prediction for row B of Lillgrund is shown in Figure \ref{fig:superposition_field}.

\section{Conclusion and Future Work}
\label{finale}
This work proposed a deep GraphSAGE neural network with jumping knowledge and residual connections (GraphSAGE+JK+Res) that is flexible and can operate directly on unstructured meshes with varying resolution. As the first attempt to introduce graph representation learning into wind turbine wake modelling, the trained GraphSAGE neural network was capable of accurately learning the complex nonlinear relationship between the inlet conditions and the resulting flow fields and achieved high prediction accuracy on data unseen during training (99.71$\%$ accuracy on predicting $U$ and 98.17$\%$ on TKE). The proposed model and workflow are highly generic and as currently formulated can be readily applied to any steady state CFD simulation on arbitrary meshes. A case study on Sweden's Lillgrund offshore wind farm was carried out using both the GAD-RANS-based CFD simulation and the proposed machine learning surrogate model. The results showed that the proposed model could accurately predict generated power compared to CFD simulation results as well as real world measured data.

With the ability to make accurate predictions under different inflow conditions and turbine yaw angles in less than 15 milliseconds, the graph neural network approach has the potential to be utilised in wind farm control and optimisation problems including yaw angle based wake steering. Future work could also include using higher fidelity LES based dynamic training data and machine learning based modelling of wake interactions.


\section*{Acknowledgements}
The authors would like to thank Dr Xiaorong Li for granting access to the GAD-CFD repository, Professor Per-Åge Krogstad of NTNU for providing access to an electronic version of the ``blind test'' dataset, and Che Liu and Dr Sibo Cheng for insightful discussions. 

\bibliographystyle{unsrt}
{\footnotesize\bibliography{citations.bib}}

\end{document}